\documentclass[conference]{IEEEtran}

\usepackage{blindtext}
\usepackage{graphicx}
\usepackage{epstopdf}

\usepackage{multirow}
\usepackage{cite}
\usepackage[ruled,vlined,commentsnumbered,linesnumbered,lined,boxed]{algorithm2e}
\usepackage{algorithmic}
\usepackage{amsmath}

\usepackage{comment}

\begin{document}

%
\title{\vspace{0.25in}A Comprehensive Study on Torchvision Pre-trained Models for Fine-grained Inter-species Classification}
\author{Feras Albardi$^1$, H M Dipu Kabir$^2$, Md Mahbub Islam Bhuiyan$^3$,\\ Parham M. Kebria$^2$, Abbas Khosravi$^2$,  Saeid Nahavandi$^{2,4}$.  \\
\IEEEauthorblockA{
$^1$King AbdulAziz University, Saudi Arabia. 
$^2$IISRI, Deakin University, Australia.
$^3$Smartify Pty Ltd, Australia.\\
$^4$ Harvard Paulson School of Engineering and Applied Sciences, Harvard University, Allston, MA 02134 USA.\\
feras.rb@gmail.com, \{hussain.kabir, abbas.khosravi, saeid.nahavandi\}@deakin.edu.au}



}


%


\maketitle

\begin{abstract}
This study aims to explore different pre-trained models offered in the Torchvision package which is available in the PyTorch library. And investigate their effectiveness on fine-grained images classification. Transfer Learning is an effective method of achieving extremely good performance with insufficient training data. In many real-world situations, people cannot collect sufficient data required to train a deep neural network model efficiently. Transfer Learning models are pre-trained on a large data set, and can bring a good performance on smaller datasets with significantly lower training time. Torchvision package offers us many models to apply the Transfer Learning on smaller datasets. Therefore, researchers may need a guideline for the selection of a good model. We investigate Torchvision pre-trained models on four different data sets: 10 Monkey Species, 225 Bird Species, Fruits 360, and Oxford 102 Flowers. These data sets have images of different resolutions, class numbers, and different achievable accuracies. We also apply their usual fully-connected layer and the Spinal fully-connected layer to investigate the effectiveness of SpinalNet. The Spinal fully-connected layer brings better performance in most situations. We apply the same augmentation for different models for the same data set for a fair comparison. This paper may help future Computer Vision researchers in choosing a proper Transfer Learning model.
\end{abstract}


\IEEEpeerreviewmaketitle

\section{Introduction}
There exist various Machine Learning (ML) libraries that made it easier to develop and implement Neural Network (NN) models \cite{qazani2020prepositioning, hossain2017application}. One of these libraries is Pytorch\cite{paszke2019pytorch} which is an open-source framework used for the research, development, and deployment of ML systems. This library includes various packages to process data and to configure models according to requirements. One of these packages is Torchvision which consists of various data sets, pre-trained models, and image processing tools and techniques. Torchvision includes many pre-trained models for Transfer Learning (TL). TL is becoming more popular as it enables for the development of high-performance Neural Network (NN) models with low data volume.

Transfer Learning is an efficient technique of using previously acquired knowledge and skills in novel problems. It is also similar to educating humans with a much broader syllabus to achieve competencies for an unpredictable future. Deep Neural Networks (DNNs) requires adequate training samples for proper training. Insufficient training samples may result in poor performance \cite{kabir2018neural, qazani2020performance, kabir2021uncertainty}. TL is an efficient DNN training technique where initial layers of DNN are pre-trained with a large data set\cite{qazani2019model}. The corresponding train data set trains only a few final layers. As a result, the user can get a well-trained NN for the specific data set of a small sample number, with lower computational overhead. TL is gaining huge popularity these days due to exceptional performance. Many researchers expect TL as the next driver of the commercial success of Machine Learning.

There are many pre-trained models in the Torchvision package. No model is good for all problems \cite{asadi2019model}. A user may not choose a model by considering its performance on the ImageNet\cite{imagenet_cvpr09} data set. Moreover, the user may have some limitation of resources and he/ she may choose a simpler low-performance model instead of a computationally demanding one. 



\section{PyTorch Pre-trained Models}
In Torchvision, there are a total of twelve main pre-trained models that are to be used for the classification of images from the fine-grained inter-species data sets. And these are AlexNet\cite{krizhevsky2014one}, VGG\cite{simonyan2014very}, ResNet\cite{he2016deep}, SqueezeNet\cite{iandola2016squeezenet}, DenseNet\cite{huang2017densely}, Inception v3\cite{szegedy2016rethinking}, GoogLeNet\cite{szegedy2015going}, ShuffleNet v2\cite{ma2018shufflenet}, MobileNet v2\cite{sandler2018mobilenetv2}, ResNeXt\cite{xie2017aggregated}, Wide ResNet\cite{zagoruyko2016wide} and MNASNet\cite{tan2019mnasnet}. Some models include multiple varieties such as ResNet and VGG \cite{kabir2021optimal, abdar2021review, kabir2019partial}. In this section, we explore these different networks and define what differentiate each one from another. 

\subsection{AlexNet}
Alex Krizhevsky introduced the novel convolutional neural network (CNN) named AlexNet in 2014. A year after competing in the ImageNet Large Scale Visual Recognition Challenge (ILSVRC)\cite{russakovsky2015imagenet} in 2012. This challenge requires the classification of a subset of ImageNet with over one million images and a thousand classes. AlexNet performed very well as it was announced the winner of the best performance for that year with top-5 error of 15.3\%, 10.8\% lower than the previous top performer. AlexNet is considered to be the first CNN to be announced the winner for the competition.

AlexNet architecture consists of a total of 8 layers, 5 convolutional layers, and 3 max-pooling layers. This architecture had many characteristics that made it perform very well, one is the utilization of ReLU (Rectified Linear Unit)\cite{nair2010rectified} activation function within the convolutional layers in replacement of the popular Tanh at the time. ReLU offered better performance and lower computation time. AlexNet benefited as well from the training on multiple graphic processors which allowed to distribute work on different GPUs enabling for the training of a bigger model and lowering the computation time. AlexNet suffered from overfitting which is a general issue that many models suffer from. Overfitting occurs when the model is fitted too well to the data and is not able to generalize well to new data which means that the model do not classify new images very good. To solve this issue, AlexNet utilized some common solutions to the problem, data augmentation, and dropout layers\cite{srivastava2014dropout}. 


\subsection{VGG}
VGG (Visual Geometry Group) is a very popular CNN that was introduced in 2014 by Karen Simonyan and Andrew Zisserman from the University of Oxford. And similar to AlexNet, VGG participated in the ILSVRC challenge in 2014. VGG scored very well as it was announced the winner (2nd place) for that year with an accuracy of 97.7\%. This architecture was able to surpass the performance of AlexNet as it had a deeper model design as well as replacing the large convolution kernels of AlexNet with multiple small convolution kernels of size 3x3.



\subsection{ResNet}
Residual Neural Network abbreviated (ResNet) is a powerful network that was proposed in 2015 by Kaiming He et al. in their paper\cite{he2016deep}. ResNet made a very big impact on deep learning as its unique architectural design enabled it to go deeper with layers and provide great performance. It was not an easy matter to get good performance with very deep models as there has been the problem of vanishing gradient. Vanishing gradient is a major problem in machine learning, it occurs as more layers are added to the network. Each layer has a certain activation function, and when the gradient is back-propagated to the earlier layers multiple times this causes the gradient to approach to zero (vanish) which makes it very hard to update the weights with very small values, and thus, the model do not converge and train well. ResNet offered a solution to the problem, in their paper, the authors described the use of a unique neural network layer named the 'Residual Block'. These layers utilize 'skip connections' which are specific shortcuts to jump over some layers. These jumps usually skip over two or three layers, and these skip connections consists of a ReLU activation function and a batch normalization in between. The main intuition behind the utilization of this structure is to avoid vanishing gradients by the re-use of activation values from a previous layer to enable the neighboring layer to learn its weight.


\subsection{SqueezeNet}

SqueezeNet has three main goals, first is to get more efficient in distributed training. Second is to export models smaller in size that are more convenient for professional and industrial clients. Lastly is to provide a model that is better deployed on embedded systems. SqueezeNet has two varieties, SqueezeNet\_v1.0 and SqueezeNet\_v1.1.

\subsection{DenseNet}
DenseNet was introduced in 2017 by Gao Huang et al. It is a powerful CNN that is similar to ResNet in using shortcut connections in between layers to solve the gradient descent problem. DenseNet utilizes 'Dense Connections' in between layers with the use of 'Dense Blocks' which holds an n number of 'Dense Layers'. Each Dense Layer consists of a 1x1 convolution filter for feature extraction and a 3x3 convolution filter to decrease the number of channels. And in Dense Blocks, each layer receives feature maps from all previous layers, and then it passes the output which is concatenated as input to all subsequent layers. This special structure allows for the use of fewer layers and the re-use of the features learned by the network, and having a narrower model (fewer layers) is easier to train with having fewer parameters to learn.


\subsection{GoogLeNet And Inception v3}
Inception\cite{szegedy2016rethinking}\cite{szegedy2015going}\cite{szegedy2016inception} is a family of CNNs developed by Christian Szegedy et al. and Google with four different versions, Inception v1 which is the GoogLeNet, Inception v2, a refined version of Inception v1 with the introduction of batch normalization, Inception v3, a more refined version with additional convolution factorization, and Inception v4.

GoogLeNet (Inception v1) was the first member of the Inception family. This architecture was introduced in 2015 and prior to that the architecture participated in the ILSVRC challenge and did very well as it won 1st place and achieved a new State-of-the-Art for ImageNet in that year. GoogLeNet was proposed for the main problem of overfitting and the expensive computation of stacking too many layers. The main solution provided by the authors is to have convolution filters of various sizes (1x1, 3x3, and 5x5) at the same level. Having these different-size filters helps in extracting features of various details, big and small, and adding more filters make the network wider rather than deeper. 
GoogLeNet architecture consists of 'Inception Modules' that carries the essence of the Inception family, and it is what allows to have convolution filters of various sizes rather than one. The idea is to let the model decide on which filter to use rather than the manual selection of a filter that might not perform well. This works with the parallel use of different filters on the same input with the same padding, and then concatenating the feature map from each filter to create one-big feature map which to be passed as input to the next inception module. 

Inception v3 was proposed in 2016. And similar to GoogLeNet it participated in the ILSVRC challenge in 2015 and scored 1st place for that year. Inception v3 holds a very similar structure to that of GoogLeNet, however, it does have some differences that enable it to generally perform better than GoogLeNet. Inception v3 architecture hold different changes, first is special convolution filter techniques (Factorized Convolutions, Smaller Convolutions, and Asymmetric Convolutions). second, a modified version of 'Auxiliary Classifiers'. Third is the use of 'Efficient Grid Size Reduction'. And lastly implementing 'Model Regularization via Label Smoothing. Factorized convolutions are utilized to reduce the number of parameters, whilst keeping the network efficient. 

\subsection{ShuffleNet v2}
Ningning Ma et al. proposed ShuffleNet v2, a CNN based on the previous ShuffleNet v1 which aimed to build an efficient and effective network with lower computation needs than other networks. With ShuffleNet v2, the authors aimed to design a network that takes into consideration the direct metrics, such as speed and memory access cost, to measure the networks computational complexity rather than an indirect metric, such as FLOPs (Floating Point Operations). FLOPs are the default measure of the performance of a model, and specifically they are units to measure how many operations are needed to run a single instance of a model. 

\subsection{MobileNet v2}
MobileNet is a family of CNNs designed to work effectively and efficiently on mobile systems. MobileNet v2 was introduced in 2018 and it is based on the previous MobileNet v1\cite{howard2017mobilenets} which was introduced in 2017. MobileNet v1 had two main layers forming the network, first is a 'Depth-wise Convolution' layer which applies a single convolution filter per input channel for lightweight filtering, and the other is a 'Point-wise Convolution' layer which consists of a 1x1 convolution filter aimed for computing linear combinations of input channels to extract new features, each of these two layers has non-linearity (ReLU).



\subsection{ResNeXt}
ResNeXt was proposed in a paper in 2017, a year after achieving 1st runner-up in the ILSVRC challenge. ResNeXt is based upon ResNet, and it differs by adding a 'NeXt' structure. This structure is a 'Cardinality' dimension which sticks on top of the width and depth of the ResNet architecture. This dimension refers to the size of the transformations set. ResNeXt is special as it replaces the use of regular linear functions in the neurons of the layers with using a non-linear function which is a part of a 'ResNeXt Block', the number of transformations in this block is referred to by the Cardinality, and after applying the required number of transformations, the results are aggregated together.


\subsection{Wide ResNet}
Wide ResNet abbreviated (WRN) is a CNN proposed in 2016 by Sergey Zagoruyko and Nikos Komodakis. It is based on the concept of residual blocks introduced in the ResNet architecture. However, it aims to be shallower with fewer layers and less training time. Wide ResNet was proposed in efforts to solve certain limitations with the ResNet architecture. ResNet has an advantage in improving performance with a large number of layers, however, adding more layers causes the reuse of diminishing feature, this affects the network by making it slow to train. The authors of WRN proposed a wider architecture (has a wider ResNet block) and lower in-depth as well as having dropout layers in between layers of residual blocks. Wider and lower in-depth means increasing the number of channel dimensions, and decreasing the number of layers.



\subsection{MNASNet}
MNASNet is a novel architecture proposed in 2018 by Mingxing Tan et al. Authors aimed to introduce a network that performs well with mobile devices as they require models that are smaller in size, yet perform effectively and efficiently. In their paper, the authors proposed three main ideas for their model, first is the introduction of a multi-objective architecture based on reinforcement learning which is used to automatically find the most suitable CNN for high performance. Second is the introduction of a novel 'Factorized Hierarchical Search Space' structure which is used as blocks in between layers and it is aimed for more efficient use of computational resources. lastly, the authors demonstrated the effectiveness of their model with regards to other architectures aimed for mobile-use, such as MobileNet, ShuffleNet, and SqueezeNet.

\begin{table}
		\caption{Number of Training , Validation and Test Examples in Each Dataset, and the Size of Images.}
		\label{Data} \centering
\begin{tabular}{|c|c|c|c|c|}	\hline
    Dataset& \multicolumn{3}{|c|}{No. of Images} & Image\\ \cline{2-4}
	 & Training  & Validation    &Test 	&  Size \\ \hline
	   
	10 Monkey Species & 1370	&	272   & -   & 400 $\times$ 300    \\ \hline
	
	Fruits 360 & 67692	&	22688   & -   & 100 $\times$ 100    \\ \hline
	225 Bird Species & 31316 & 1125 & 1125 & 224 $\times$ 224    \\ \hline
	Oxford 102 Flower & 1020 & 1020 & 6149 & Varying    \\ \hline
\end{tabular} 
\end{table}

\begin{table}
		\caption{Number of Layers and Parameters in Torchvision Pre-trained Models}
		\label{Model} \centering
\begin{tabular}{|c|c|c|c|c|}	\hline 

    Model & No. of  &No. of      &Year of 	& Top-1 error \\ 
          &  Layers & Parameters & Proposal & on Imagenet \\
    
    \hline AlexNet & 8	   &    61M    & 2014  & 43.45\%   \\ 
    \hline VGG-11  & 11    &   132M    & 2014  & 30.98\%   \\ 
    \hline VGG-13  & 13    &   133M    & 2014  & 30.07\%   \\
    \hline VGG-16  & 16    &   138M    & 2014  & 28.41\%   \\
    \hline VGG-19  & 19    &   143M    & 2014  & 27.62\%   \\
    \hline VGG-11 BN    & 11    & 132M   & 2014  & 29.62\% \\
    \hline VGG-13 BN    & 13    & 133M   & 2014  & 28.45\% \\
    \hline VGG-16 BN    & 16    & 138M   & 2014  & 26.63\% \\
    \hline VGG-19 BN    & 19    & 143M   & 2014  & 25.76\% \\
    \hline ResNet-18    & 18    & 11M    & 2015  & 30.24\% \\
    \hline ResNet-34    & 34    & 21M    & 2015  & 26.70\% \\
    \hline ResNet-50    & 50    & 25M    & 2015  & 23.85\% \\
    \hline ResNet-101   & 101   & 44M    & 2015  & 22.63\% \\
    \hline ResNet-152   & 152   & 50M    & 2015  & 21.69\% \\
    \hline SqueezeNet 1.0 &  -  & 1M     & 2017  & 41.90\% \\
    \hline SqueezeNet 1.1 &  -  & 1M     & 2017  & 41.81\% \\
    \hline DenseNet-121 & 121   & 7M     & 2016  & 25.35\& \\
    \hline DenseNet-169 & 169   & 14M    & 2016  & 24.00\% \\
    \hline DenseNet-161 & 161   & 28M    & 2016  & 22.35\% \\
    \hline DenseNet-201 & 201   & 20M    & 2016  & 22.80\% \\
    \hline Inception v3 & 48    & 27M    & 2015  & 22.55\% \\
    \hline GoogLeNet    & 22    & 6M     & 2014  & 30.22\% \\
    \hline ShuffleNet v2 &  -   & 2M     & 2018  & 30.64\% \\
    \hline MobileNet v2  &  53  & 4M     & 2018  & 28.12\% \\
    \hline ResNeXt-50   & 50    & 25M    & 2016  & 22.38\% \\
    \hline ResNeXt-101  & 101   & 88M    & 2016  & 20.69\% \\
    \hline Wide ResNet-50-2 & 50 & 68M   & 2016  & 21.49\% \\
    \hline Wide ResNet-101-2 & 101 & 126M & 2016 & 21.16\% \\
    \hline MNASNet 1.0 & -         & 4M   & 2018 & 26.49\% \\
    \hline
\end{tabular} 
\end{table}


\section{Methodology}
This study was performed in several systematic steps that generalizes over different data sets and models to ensure consistent results.

\subsection{Phase 1. Data Collection and Preprocessing}
There are four different sets of fine-grained Inter-species data. The first one, named 10 Monkey Species\cite{mario2018monkey}, is a set which includes a total of 1400 images of size 400x300 distributed among 10 different species of monkeys. These species include Alouattapalliata, Erythrocebuspatas, Cacajaocalvus, Macacafuscata, Cebuellapygmea, Cebuscapucinus, Micoargentatus, Saimirisciureus, Aotusnigriceps and Trachypithecusjohnii. The collection of these images was done with the help of search engine web scrapers. 10 Monkey Species is available on the popular data platform, Kaggle.

\begin{figure}
  \centering
  \includegraphics[width=3.2in,angle=0]{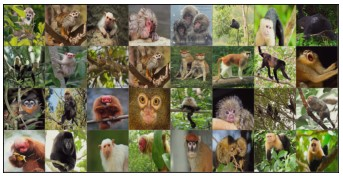}
  \caption{A batch of 32 samples from the 10 Monkey Species data set.}
  \label{Mon_N}
\end{figure}

The second set named Fruits 360\cite{murecsan2018fruit}. It includes data of different fruits and vegetables. This set is much bigger than 10 Money Species with more than 90,000 images of fruits and vegetables of size 100x100 and ranging from watermelons, apples and oranges to potatoes and onions and much more with 103 classes. 

\begin{figure}
  \centering
  \includegraphics[width=3.2in,angle=0]{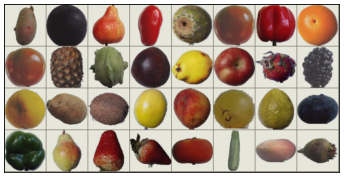}
  \caption{A batch of 32 samples from the Fruits 360 data set.}
  \label{Fruits_N}
\end{figure}

225 Bird Species\cite{gerry2020bird} is the third data set. It includes more than 30,000 training, testing, and validation images distributed among 225 classes of birds, all of these images are of size 224x224.

\begin{figure}
  \centering
  \includegraphics[width=3.2in,angle=0]{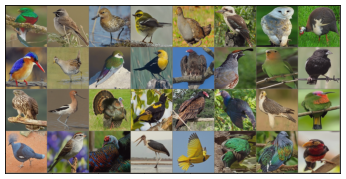}
  \caption{A batch of 32 samples from the 225 Bird Species data set.}
  \label{Bird_N}
\end{figure}

Oxford 102\cite{Nilsback08} is the fourth and last data set, and it consists of 102 classes of different flowers. This data set was created by the Visual Geometry Group of Oxford. The type of flowers available in the data set are commonly available in the United Kingdom. And each class holds between 40 and 258 images. Collection of images was done using web scrappers as well as manually capturing flower images.

\begin{figure}
  \centering
  \includegraphics[width=3.2in,angle=0]{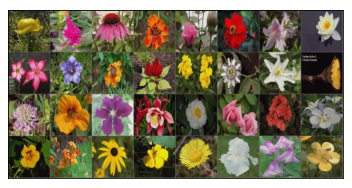}
  \caption{A batch of 32 samples from the Oxford 102 flower data set.}
  \label{Flower_N}
\end{figure}

One of the methods to increase the performance of models is to provide more training examples \cite{qazani2020new, pedrammehr2018novel}. this method can be accomplished with data augmentation where random examples are taken from the data set, and then apply different image processing methods on these examples, such as rotation, horizontal mirroring, and magnification. This allows for the addition of many examples to the training set. The main objective of utilizing image augmentation techniques is its proven effect on improving performance \cite{qazani2019high, qazani2019linear}. Having a set with more images means that the model is faced with a bigger number of samples that can highly improve performance especially with low-volume data sets.

\begin{figure}
  \centering
  \includegraphics[width=3.2in,angle=0]{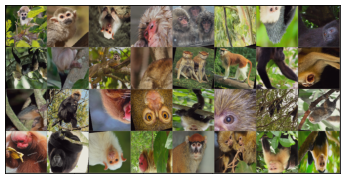}
  \caption{A batch of 32 augmented samples from the 10 Monkey Species data set.}
  \label{Mon_A_A}
\end{figure}

\begin{figure}
  \centering
  \includegraphics[width=3.2in,angle=0]{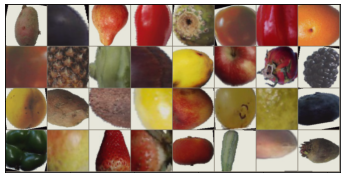}
  \caption{A batch of 32 augmented samples from the Fruits 360 data set.}
  \label{Bird_N_A}
\end{figure}

\begin{figure}
  \centering
  \includegraphics[width=3.2in,angle=0]{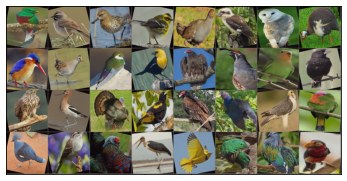}
  \caption{A batch of 32 augmented samples from the 225 Bird Species data set.}
  \label{Bird_A_A}
\end{figure}

\begin{figure}
  \centering
  \includegraphics[width=3.2in,angle=0]{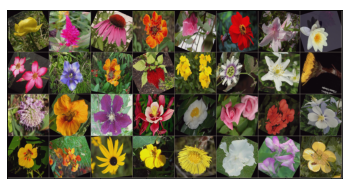}
  \caption{A batch of 32 augmented samples from the Oxford 102 flower data set.}
  \label{Flower_A_A}
\end{figure}

\subsection{Phase 2. Models Training}
In this phase, we train each data set independently with 29 different pre-trained models available in the Torchvision package. These models include AlexNet, VGG (11, 13 16, 19, BN-11, BN-13, BN-16, BN-19), ResNet (18, 34, 50, 101, 152), SqueezeNet (1.0, 1.1), DenseNet (121, 169, 161, 201), Inception v3, GoogLeNet, ShuffleNet v2 (0.5, 1.0), MobileNet v2, ResNeXt (50, 101), Wide ResNet (50, 101) and MNASNet (1.0).
For the training and development environment, we utilized Kaggle Kernels which are in essence Jupyter\cite{soton403913} Notebooks as they offer ease of importing required datasets.

\subsection{Phase 3. Investigating Results}
This phase includes the extraction of results and performance of each model with each data set and form it all together in tables in order to understand and compare how each model performs with fine-grained Inter-species data. We investigate the results to find the best possible configurations for these data with Torchvision pre-trained models, and to suggest new information that can possibly lead for the implementation of more performant models.

\subsection{Phase 4. Investigating SpinalNet}
This last phase is actually a derivative of the third phase. Here we implement each data set with pre-trained models with a small configuration. This change is to replace the last fully connected layer in each model with the SpinalNet\cite{kabir2020spinalnet} layer. SpinalNet is a novel network that is inspired from the human somatosensory system \cite{abbas2020spinenet}. This network consists of input row, intermediate row and output row where the intermediate row contains a small number of neurons, and each hidden layer receives a part of the input and outputs of the previous layer with input segmentation.


\section{Results}
Exploring the results of the first data set, 10 Monkey Species, it can be noticed that pre-trained models provided very good performance with regards to the low data volume. Training hyperparameters are a batch of 40 images (20 for DenseNet-169, DenseNet-161, DenseNet-201, and ResNeXt-101. This is due to memory limitations.), 10 epochs, SGD optimizer, and varying values of the learning rate which depends on the model in-hand. Most models provided above 90\% accuracy, while others, such as SqueezeNet 1.0 ended with an accuracy of 65.80\% (Normal FC) and 69.11\% (Spinal FC). However, SqueezeNet 1.1 did provide a better accuracy with about 84.55\% (Normal FC) and 82.35\% (Spinal FC). It is critical to note that models which performed poorly, such as SqueezeNet and MNASNet, were able to perform much better when trained using the Adam\cite{kingma2014adam} optimizer instead of SGD, and that goes for both Normal and Spinal FC. However, it was decided to move forward with SGD results as to be consistent with all other models. And at the end of this section (in figure 9 and 10), we illustrated the accuracy and loss over epochs for the Wide ResNet-101-2 model which was selected randomly to view performance with both Normal and Spinal FC layer.

\begin{table}
		\caption{Performance of pre-trained models with the 10 Monkey Species data set with both default and SpinalNet FC}
		\label{Monkey_Results} \centering
\begin{tabular}{|c|c|c|c|}	\hline 

    Model & Batch   & Accuracy    & Accuracy    \\ 
          & Size     & (Normal FC) & (Spinal FC) \\
    
    \hline AlexNet & 40 &   90.80\%    & 90.44\% \\
    \hline VGG-11  & 40 &   96.69\%    & 96.69\%   \\ 
    \hline VGG-13  & 40 &   97.43\%    & 97.43\%   \\
    \hline VGG-16  & 40 &   96.69\%    & 97.43\%   \\
    \hline VGG-19  & 40 &   97.43\%    & 97.79\%   \\
    \hline VGG-11 BN    & 40 & 95.59\% & 97.43\%   \\
    \hline VGG-13 BN    & 40 & 98.16\% & 98.53\% \\
    \hline VGG-16 BN    & 40 & 98.53\% & 98.53\% \\
    \hline VGG-19 BN    & 40 & \textbf{98.90\%} & 98.90\% \\
    \hline ResNet-18    & 40 & 94.85\%  & 98.16\% \\
    \hline ResNet-34    & 40 & 97.05\%  & 98.89\% \\
    \hline ResNet-50    & 40 & 97.05\%  & 98.52\% \\
    \hline ResNet-101   & 40 & 96.32\%  & \textbf{99.26\%} \\
    \hline ResNet-152   & 40 & 96.32\%  & 98.16\% \\
    \hline SqueezeNet 1.0 & 40 & 65.80\% & 69.11\% \\
    \hline SqueezeNet 1.1 & 40 & 84.55\% & 82.35\% \\
    \hline DenseNet-121 & 40 & 95.95\%  & 95.95\% \\
    \hline DenseNet-169 & 20 & 95.58\%  & 97.79\% \\
    \hline DenseNet-161 & 20 & 93.75\%  & 92.64\% \\
    \hline DenseNet-201 & 20 & 96.69\%  & 97.05\% \\
    \hline Inception v3 & 40 & 97.79\%  & \textbf{99.26\%} \\
    \hline GoogLeNet    & 40 & 97.05\%  & 98.16\% \\
    \hline ShuffleNet v2 0.5 & 40  & 91.91\%   & 93.01\% \\
    \hline ShuffleNet v2 1.0 & 40  & 96.32\%   & 95.95\% \\
    \hline MobileNet v2 & 40 & 94.85\%  & 94.11\% \\
    \hline ResNeXt-50   & 40 & 98.16\%  & 98.89\% \\
    \hline ResNeXt-101  & 20 & 97.79\%  & 98.52\% \\
    \hline Wide ResNet-50-2  & 40 & 98.16\% & 98.53\% \\
    \hline Wide ResNet-101-2 & 40 & 98.53\% & \textbf{99.26\%} \\
    \hline MNASNet 1.0       & 40 & 79.05\% & 93.01\% \\
    \hline
\end{tabular} 
\end{table}

In Fruits 360, training was performed on a batch size of 224 (112 for DenseNet-169, DenseNet-161, DenseNet-201, and ResNeXt-101. This is due to memory limitations.), and a default selection of 10 epochs, SGD optimizer, and varying values of the learning rate in accordance with different models. Almost all Models provided great performance reaching to an accuracy of above 99\% with only 10 epochs. This is mostly due to the fact that this data set has a high number of examples. The only model which performed poorly is MNASNet with an accuracy of 58.51\% (Normal FC) and 61.17\% (Spinal FC). And similar to the previous data set, it was found that MNASNet is able to provide great performance similar to that of other models in the list when using the Adam optimizer. However, and for consistency we moved forward with SGD results. And it is worth noting that the best performance was provided by the DenseNet169 model with the SpinalNet fully connected layer with an accuracy of 100.0\%.

\begin{table}
		\caption{Performance of pre-trained models with the Fruits 360 data set with both default and SpinalNet FC}
		\label{Fruits_Results} \centering
\begin{tabular}{|c|c|c|c|}	\hline 

    Model & Batch   & Accuracy    & Accuracy    \\ 
          & Size    & (Normal FC) & (Spinal FC) \\
    
    \hline AlexNet & 224    &   96.64\%    & 96.09\% \\
    \hline VGG-11  & 224    &   99.78\%    & 99.82\%   \\ 
    \hline VGG-13  & 224    &   99.85\%    & 99.90\%   \\
    \hline VGG-16  & 224    &   99.84\%    & 99.90\%   \\
    \hline VGG-19  & 224    &   99.86\%    & 99.90\%   \\
    \hline VGG-11 BN    & 224    & 99.78\% & 99.85\%   \\
    \hline VGG-13 BN    & 224    & 99.79\% & 99.88\% \\
    \hline VGG-16 BN    & 224    & 99.88\% & 99.93\% \\
    \hline VGG-19 BN    & 224    & 99.90\% & 99.96\% \\
    \hline ResNet-18    & 224    & 99.81\% & 99.66\% \\
    \hline ResNet-34    & 224    & 99.84\% & 99.71\% \\
    \hline ResNet-50    & 224    & 99.93\% & 99.78\% \\
    \hline ResNet-101   & 224    & 99.97\% & 99.87\% \\
    \hline ResNet-152   & 224    & \textbf{99.98\%}  & 99.81\% \\
    \hline SqueezeNet 1.0 & 224  & 99.02\% & 98.31\% \\
    \hline SqueezeNet 1.1 & 224  & 99.25\% & 98.77\% \\
    \hline DenseNet-121   & 224  & 99.68\% & 99.75\% \\
    \hline DenseNet-169 & 112 & \textbf{99.98\%} & \textbf{100\%} \\
    \hline DenseNet-161 & 112 & 99.83\%  & 99.93\% \\
    \hline DenseNet-201 & 112 & 99.78\%  & 99.94\% \\
    \hline Inception v3 & 224 & 99.92\%  & 99.60\% \\
    \hline GoogLeNet    & 224 & 99.87\%  & 98.94\% \\
    \hline ShuffleNet v2 0.5  & 224 & 98.39\%   & 98.68\% \\
    \hline ShuffleNet v2 1.0  & 224 & 99.36\%   & 99.78\% \\
    \hline MobileNet v2 & 224 & 99.51\%  & 99.91\% \\
    \hline ResNeXt-50   & 224 & 99.86\%  & 99.89\% \\
    \hline ResNeXt-101  & 112 & \textbf{99.98\%}& 99.90\% \\
    \hline Wide ResNet-50-2   & 224 & 99.96\%   & 99.96\% \\
    \hline Wide ResNet-101-2  & 224 & 99.96\%   & \textbf{100\%} \\
    \hline MNASNet 1.0        & 224 & 58.51\%   & 61.17\% \\
    \hline
\end{tabular} 
\end{table}

225 Bird Species provided very good performance similar to that of Fruits 360. With configured hyperparameters of a batch size of 64 (32 for DenseNet-169, DenseNet-161, DenseNet-201, and ResNeXt-101. This is due to memory limitations.), 10 epochs, SGD optimizer, and varying values of the learning rate which depends on the model in-hand. Most models performed an accuracy of above 90\%, such as VGG-16 99.11\% (Normal FC) and 99.20\% (Spinal FC) as well as the best performer the Wide ResNet-101-2 with 99.38\% (Normal FC) and 99.56\% (Spinal FC). Lowest performance was achieved with SqueezeNet 1.0 and SqueezeNet 1.1 with an accuracy of 82.95\% (Normal FC), 84.61\% (Spinal FC) and 99.08\% (Normal FC), 84.17\% (Spinal FC) respectively. 

\begin{table}
		\caption{Performance of pre-trained models with the 225 Bird Species data set with both default and SpinalNet FC}
		\label{Bird_Results} \centering
\begin{tabular}{|c|c|c|c|}	\hline 

    Model & Batch   & Accuracy    & Accuracy       \\ 
          & Size    & (Normal FC) & (Spinal FC)    \\
    
    \hline AlexNet & 64 &   93.33\%    & 93.24\%   \\      
    \hline VGG-11  & 64 &   98.87\%    & 99.02\%   \\ 
    \hline VGG-13  & 64 &   99.06\%    & 99.11\%   \\
    \hline VGG-16  & 64 &   99.11\%    & 99.20\%   \\
    \hline VGG-19  & 64 &   99.16\%    & 99.20\%   \\
    \hline VGG-11 BN    & 64   & 99.02\% & 99.02\% \\
    \hline VGG-13 BN    & 64   & 99.11\% & 99.19\% \\
    \hline VGG-16 BN    & 64   & 99.08\% & 99.29\% \\
    \hline VGG-19 BN    & 64   & 99.02\% & 99.11\% \\
    \hline ResNet-18    & 64   & 98.48\% & 98.48\% \\
    \hline ResNet-34    & 64   & 98.13\% & 98.66\% \\
    \hline ResNet-50    & 64   & 98.66\% & 98.66\% \\
    \hline ResNet-101   & 64   & 98.75\% & 98.93\% \\
    \hline ResNet-152   & 64   & 98.75\% & 99.02\% \\
    \hline SqueezeNet 1.0 & 64 & 82.95\% & 84.61\% \\    
    \hline SqueezeNet 1.1 & 64 & 88.08\% & 84.17\% \\
    \hline DenseNet-121 & 64   & 98.75\% & 99.11\% \\
    \hline DenseNet-169 & 32   & 98.57\% & 98.93\%  \\
    \hline DenseNet-161 & 32   & 99.02\% & 98.93\% \\ 
    \hline DenseNet-201 & 32   & 98.75\% & 98.84\% \\
    \hline Inception v3 & 64   & 97.95\% & 98.57\% \\
    \hline GoogLeNet    & 64   & 98.31\% & 98.04\% \\
    \hline ShuffleNet v2 0.5   & 64 & 83.82\%  & 84.48\% \\  
    \hline ShuffleNet v2 1.0   & 64 & 94.43\%  & 93.30\% \\   
    \hline MobileNet v2 &  64  & 98.48\% & 98.75\% \\
    \hline ResNeXt-50   & 64   & 98.93\% & 98.40\% \\
    \hline ResNeXt-101  & 32   & 98.75\% & 98.84\% \\    
    \hline Wide ResNet-50-2    & 64      & 99.29\%  & 99.38\% \\
    \hline Wide ResNet-101-2   & 64      & \textbf{99.38\%}  & \textbf{99.56\%} \\
    \hline MNASNet 1.0         & 64      & 96.95\%           & 96.34\% \\
    \hline
\end{tabular} 
\end{table}

Last data set in this study is the Oxford 102 Flower data set. Most models were able to achieve an accuracy above 90\% with different variations among them. Models, such as MobileNet v2 and AlexNet, achieved an accuracy of 91.02\% and 92.78\% respectively. And models, such as DenseNet-201, performed best with 98.29\% for Normal FC and 98.36\% for Spinal FC as well as Wide ResNet-101-2 which achieved 98.29\% for both Normal and Spinal FC. It can be noticed that Spinal FC was able to averagely provide a better performance than that of Normal FC.

\begin{table}
		\caption{Performance of pre-trained models with the Oxford 102 Flower data set with both default and SpinalNet FC}
		\label{Flowers_Results} \centering
\begin{tabular}{|c|c|c|c|}	\hline 

    Model & Batch   & Accuracy    & Accuracy    \\ 
          & Size     & (Normal FC) & (Spinal FC) \\
    
    \hline AlexNet & 64    &  92.78\%   & 92.66\% \\
    \hline VGG-11  & 64    &   94.56\%    & 94.67\%   \\ 
    \hline VGG-13  & 64    &   94.56\%    & 95.01\%   \\
    \hline VGG-16  & 64    &   94.67\%    & 95.01\%   \\
    \hline VGG-19  & 64   &   95.32\%    & 95.17\%   \\
    \hline VGG-11 BN    & 64 &  94.62\%    & 94.67\%   \\
    \hline VGG-13 BN    & 64  &  94.76\%    & 95.01\%   \\
    \hline VGG-16 BN    & 64  &  94.97\%    & 95.17\%   \\
    \hline VGG-19 BN    & 64  &  95.11\%    & 95.38\%   \\
    \hline ResNet-18    & 64  &  95.83\%    & 96.14\%   \\
    \hline ResNet-34    & 64 &  96.72\%    & 96.72\%   \\
    \hline ResNet-50    & 64 &  97.43\%    & 97.55\%   \\
    \hline ResNet-101   & 64 &  97.92\%    & 97.92\%   \\
    \hline ResNet-152   & 64 &  97.92\%    & 98.04\%   \\
    \hline SqueezeNet 1.0 & 64 & 95.01\%    & 95.14\%   \\
    \hline SqueezeNet 1.1 & 64 & 95.26\%    & 95.20\%   \\ 
    \hline DenseNet-121 & 64 &  97.58\%    & 97.64\%   \\
    \hline DenseNet-169 & 32 &  98.00\%    & 98.04\%   \\
    \hline DenseNet-161 & 32 &  98.04\%    & 98.17\%   \\
    \hline DenseNet-201 & 32 &  \textbf{98.29}\%    & \textbf{98.36\%}   \\
    \hline Inception v3 & 64 &  98.17\%    & 98.19\%   \\
    \hline GoogLeNet    & 64 &  95.93\%    & 96.01\%   \\
    \hline ShuffleNet v2 0.5 & 64 &  92.78\%    & 92.66\%   \\
    \hline ShuffleNet v2 1.0 & 64 &  93.22\%    & 93.45\%   \\
    \hline MobileNet v2  & 64 &  91.02\%    & 90.94\%   \\
    \hline ResNeXt-50   & 64 &  97.80\%    & 97.80\%   \\
    \hline ResNeXt-101  & 32 &  98.00\%    & 98.04\%   \\
    \hline Wide ResNet-50-2 & 64 &  98.04\%    & 98.17\%   \\
    \hline Wide ResNet-101-2 & 64 &  \textbf{98.29}\%    & 98.29\%   \\
    \hline MNASNet 1.0 & 64 &  95.67\%    & 95.59\%   \\
    \hline
\end{tabular} 
\end{table}

SpinalNet is utilized to explore the performance of the novel network and compare it with the default network in each model. All models were trained the same. However, with SpinalNet we replace the default fully connected layer with the SpinalNet fully connected layer. SpinalNet provided very good performance as it has actually improved accuracies of most models over the default FC layer, an example would be with the 10 Monkey Species data set, MNASNet 1.0 provided an accuracy of 79.05\% accuracy, and the same model with Spinal fully connected layer achieved an accuracy of 93.01\%. And with the Oxford 102 Flower data set, VGG-13 provided an accuracy of 94.56\%, and VGG-13 with Spinal fully connected layer achieved an accuracy of 95.01\%. Although in a number of cases, the default FC layer did provide better performance. And overall, SpinalNet shows promising results in the classification of images, and it is possible to provide very good performance if integrated well with pre-trained models .

At first, we aimed to train all models with a constant set of hyperparameters. However, it turned out that several models performed much better when adjusting a specific parameter, which is the learning rate, to different values than the default one. AlexNet, SqueezeNet, and others performed moderately when trained with the default learning rate value. And We found that using Adam instead of SGD helped drastically in improving performance. However, this contradicts our goal of utilizing a specific optimizer instead of utilizing different optimizers with different models which can potentially create inconsistency. And with several attempts, we found that adjusting the learning rate when using SGD helped greatly in providing similar, and in some cases better performance, than changing the optimizer to Adam. This improvement in adjusting the value of the learning rate with different models reflects to all data sets.



\section{Conclusions}
This study aimed to investigate the performance of various pre-trained models offered in the Torchvision package of Pytorch with fine-grained inter-species datasets. Each model has a unique architecture designed to perform best with visual data. We systematically fine-tuned each model with a configured set of hyperparameters to ensure consistency of results over different models. Models in the training of data sets with high volume of examples, such as the Fruits 360, provided great performance with most models reaching to 99\% and above. This is also applicable with other data sets with a small number of examples, such as 10 Monkey Species, most models were still able to achieve very good results. We explored the novel SpinalNet architecture in efforts to confirm the improvement of pre-trained models performance. In many models, SpinalNet did perform very well reaching in some models to an accuracy of 100\% in only 10 epochs for the Fruits 360 data set. And in other models, SpinalNet did perform well and was able to improve the accuracy. However, in many models, the normal fully connected layer performed similar or better. SpinalNet shows very promising results of improving the performance of pre-trained models. There are no specific models that showed best performance with the novel network. It may depend on the number of examples in the data set to provide better performance. SpinalNet provided better accuracies with data sets consisting of a large volume of examples. 


\section*{Acknowledgment}
I (Feras) would like to thank my brother Faris for his continuous support and encouragement.

\bibliographystyle{IEEEtran}
\bibliography{Ref}

\end{document}